\documentclass[conference]{IEEEtran}
\IEEEoverridecommandlockouts
\usepackage{cite}
\usepackage{amsmath,amssymb,amsfonts}
\usepackage{algorithmic}
\usepackage{graphicx}
\usepackage{textcomp}
\usepackage{xcolor}
\usepackage{hyperref}

\ifCLASSOPTIONcompsoc
\usepackage[caption=false, font=normalsize, labelfont=sf, textfont=sf]{subfig}
\else
\usepackage[caption=false, font=footnotesize]{subfig}
\fi

\def\BibTeX{{\rm B\kern-.05em{\sc i\kern-.025em b}\kern-.08em
    T\kern-.1667em\lower.7ex\hbox{E}\kern-.125emX}}

\title{Adapted Pepper}

\begin{document}

\author{\IEEEauthorblockN{Maxime Caniot\IEEEauthorrefmark{1}, Vincent Bonnet\IEEEauthorrefmark{2}, Maxime Busy, Thierry Labaye, Michel Besombes,\\ Sebastien Courtois and Edouard Lagrue}
\IEEEauthorblockA{SoftBank Robotics Europe\\
%Email: \IEEEauthorrefmark{1}mbusy@softbankrobotics.com, \IEEEauthorrefmark{1}mcaniot@softbankrobotics.com
}
}

\maketitle
\thispagestyle{plain}
\pagestyle{plain}

%%%%%%%%%%%%%%%%%%%%%%%%%%%%%%%%%%%%%%%%%%%%%%%%%%%%%%%%%%%%%%%%%%%%%%%%%%%%%%%%
\begin{abstract}

One of the main issue in robotics is the lack of embedded computational power. Recently, state of the art algorithms providing a better understanding of the surroundings (Object detection, skeleton tracking, etc.) are requiring more and more computational power. The lack of embedded computational power is more significant in mass-produced robots because of the difficulties to follow the increasing computational requirements of state of the art algorithms. The integration of an additional GPU allows to overcome this lack of embedded computational power.
We introduce in this paper a prototype of Pepper with an embedded GPU, but also with an additional 3D camera on the head of the robot and plugged to the late GPU. This prototype, called Adapted Pepper, was built for the European project called MuMMER (MultiModal Mall Entertainment Robot) in order to embed algorithms like OpenPose, YOLO or to process sensors information and, in all cases, avoid network dependency for deported computation.

\end{abstract}

\bigskip


%%%%%%%%%%%%%%%%%%%%%%%%%%%%%%%%%%%%%%%%%%%%%%%%%%%%%%%%%%%%%%%%%%%%%%%%%%%%%%%%
\section{INTRODUCTION}

Deported computation is a common way to improve and use state of the art algorithms, such as OpenPose~\cite{openpose} and YOLO~\cite{yolo}, without modifying an important part of the robot hardware. However, this solution is limited by the network interface restriction like security access, disconnection or even the bandwidth. These restrictions can cause latency or impact the accuracy of the algorithms by reducing the size of data that it is possible to send through the network (WiFi 802.11~\cite{wifi_benchmark}: from 11 to 600 Mbps, GigabitEthernet~\cite{gigabitethernet_benchmark}: 1000 Mbps).

In the field of Humanoid Robot Interaction (HRI), the necessity to have a reactive architecture is as important as the content of the interaction. A latency in the interaction diminishes the satisfaction of the user~\cite{kuhmann1987}\cite{shneiderman2016} or even influences the emotional state of the user~\cite{delay_emotion_effect}. However, a time of response of one or two seconds in the interaction with a human is suggested to be more humanlike~\cite{shiomi2017subtle}\cite{shiwa2009quickly}. Reducing the performances of the used algorithms or improving the hardware to embed the computation~\cite{reyes2018near} can reduce the latency with the benefice of a smoother interaction.

This dilemma was encountered in the European project called MuMMER~\cite{foster2016mummer} (MultiModal Mall Entertainment Robot). The goal of the project was to develop a humanoid robot able to interact autonomously and naturally in a public shopping mall.

\begin{figure}[htbp]
\centering
 \includegraphics[scale=0.20]{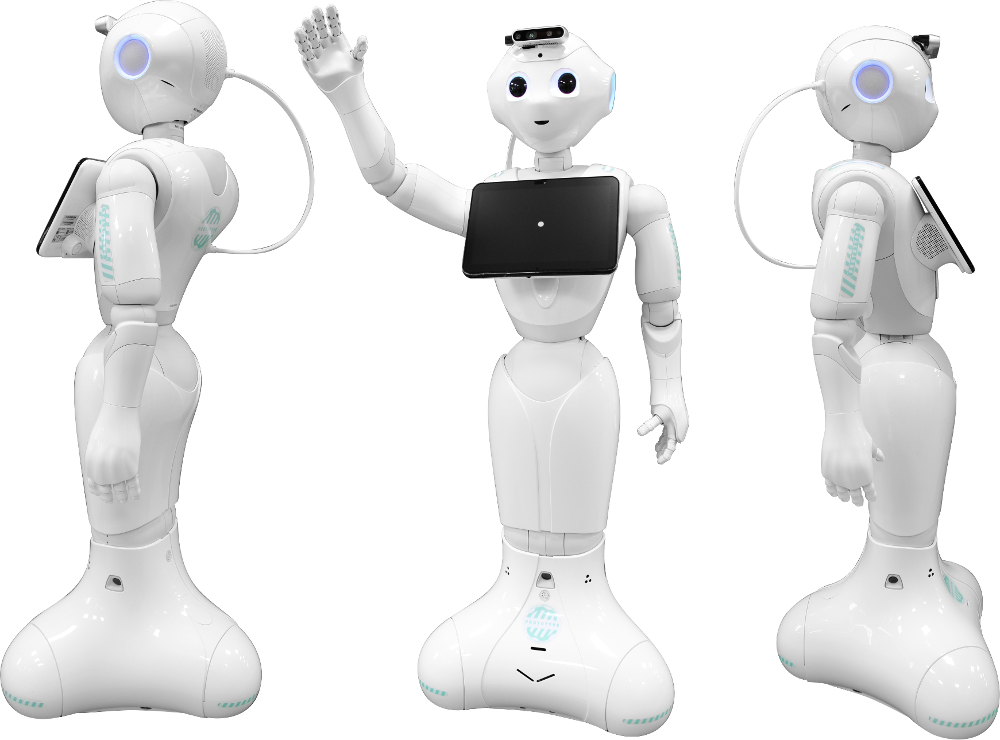}
\caption{Different views of the Adapted Pepper}
\label{fig:adapted_pepper}
\end{figure}

A prototype called Adapted Pepper(Fig.\ref{fig:adapted_pepper}) is developed to increase the quality of interaction thanks to an improvement of the hardware of Pepper.
The Adapted Pepper is a prototype based on a Pepper 1.8\footnote{\url{http://doc.aldebaran.com/2-5/family/pepper_technical/index_pep.html}} with an additional GPU and an additional 3D camera.

The integration of a GPU within Pepper has already been studied by the university of Chile~\cite{reyes2018near} and by the university of Salerno~\cite{miviabot}. These integrations had the same purpose as Adapted Pepper, to enhance the perception capabilities of the robot, specifically in the fields of:
\begin{itemize}
    \item Object recognition
    \item Face detection
    \item Age and gender recognition
    \item Emotion recognition
\end{itemize}
In \cite{reyes2018near}, the GPU was integrated as a Backpack and directly connected to Pepper through an Ethernet cable. The GPU and the camera are further integrated into the Adapted Pepper to answer to design constraints.

This paper introduces the selected devices for the integration (the Intel D435\footnote{\url{https://www.intelrealsense.com/depth-camera-d435}} and the Jetson TX2~\cite{jetsonTX2Benchmark}), the experiments conducted on each device before their integration, and the integration of these devices into the robot. Finally, the software architecture is presented and validated by running Deep Learning models on the Adapted Pepper. 

\section{Hardware}

As aforementioned, the Adapted Pepper is based on Pepper version 1.8. The integration of the camera and the GPU on the Pepper robot needs to compose with different constraints. The main constraints are :
\begin{itemize}
    \item Mechanical Constraints: the integration of the new devices must not deteriorate the mechanical functionalities of Pepper.
    \item Performance Constraints: the integration of the new devices must not deteriorate the behavior of the robot and their own functions.
    \item Temperature Constraints: the integration of the new devices must not increase the internal temperature of Pepper
    \item Electrical power Constraints: the consumption of the new devices must not severely deteriorate the autonomy of the robot.
    \item Design Constraints: The new devices must be integrated into the robot with a minimum impact on the current design of the robot.
\end{itemize}{}

\subsection{Selected devices}

\subsubsection{Depth Camera}

The Pepper 1.8 already has an embedded stereo camera but the reconstruction of the 3D images is done by the CPU of the robot. In order to avoid an overload of the CPU, the 3D reconstruction is based on fewer frames and images of lesser resolution than the stereo camera could achieve.  The 3D reconstruction quality and frame rate are insufficient to analyze the environment at close range and in real time. The 3D Camera of Pepper is designed for a mass product market. The research fields of MuMMER need a higher camera performance. Moreover, a camera with an embedded processor allows to save the corresponding computational power for the CPU. In summary, an additional 3D camera, with an embedded processor, allows to obtain high resolution images without burdening CPU capacities with costly 3D reconstruction while respecting the aforementioned Performance Constraints.
Due to the architecture of Pepper 1.8, two configurations are possible for connecting the 3D camera. First configuration, the camera is connected to the head of the robot (robot connected to the GPU itself). The images provided by the camera are sent to the robot head and afterward to the GPU. Second configuration, the camera is directly connected to the GPU. As the goal is to send the images on the GPU, the second configuration is chosen. Directly connecting this camera to the GPU avoids any latency and preserves the 3D images qualities. The position of this additional camera, on the robot forehead, is chosen to allow human tracking and preserve the design of the robot. The use-case with the 3D camera is human tracking (face, skeleton, gesture, etc.) requiring a high accuracy on a short range (0 to 1.5m). The NAOqi framework defines 3 different engagement zones\footnote{\url{http://doc.aldebaran.com/2-5/naoqi/peopleperception/alengagementzones.html}}:
\begin{itemize}
    \item Zone 1, 0 to 1.5m, also called engagement zone: zone where the robot can interact with humans (dialogue, face tracking, etc.)
    \item Zone 2, 1.5 to 2.5m, also called pre-engagement: zone where the robot can only track a human.
    \item Zone 3, 2.5m and beyond, also called non-engagement zone: zone where the robot cannot interact with humans nor track humans.
\end{itemize}{}

Based on these specifications, the camera selected for the integration is the Intel D435 camera. The Intel camera is based on active infrared (IR) stereo. It is composed of an IR projector combined with two imagers. The projector projects non-visible static IR patterns that are captured by the left and right imagers and afterwards sent to a dedicated depth imaging processor. Moreover, a RGB camera is also embedded within this device. The realsense camera D435 specifications are :
\begin{itemize}
    \item Range(m): 0.2-10
    \item Depth resolution(pixels): 1280 x 720
    \item Interface: USB-C 3.1 Gen 1
\end{itemize}{}

This camera is chosen for its accuracy in short range~\cite{camera_benchmark}. The D435 has indeed a low bias (bias of 0.05m for a distance inferior to 1.5m) and a good precision (precision of 0.05m for a distance inferior to 6m). Furthermore, this device is not affected by the noise of additional sensors. Finally, its size allows an easier integration and adaptation with our robot design. In parallel, the integration needs to take into account the thermal sensitivity of the camera\footnote{\url{https://www.mouser.com/pdfdocs/Intel_D400_Series_Datasheet.pdf}}. The IR projections diverge with heat and reduce the accuracy of the 3D images. The integration needs to take into account the heat already produced within the head of Pepper by the embedded processor of the camera and the heat produced by the top camera as well. Indeed, the position of integration of the added 3D camera is next to the original top camera of Pepper. In order to not deteriorate the D435 camera accuracy, its integration needs to dissipate the extra heat.

\subsubsection{GPU}

The current processor of Pepper 1.8 does not provide enough computational power to support Deep Learning algorithms and the behaviors of the robot in parallel. Adding a new processor with a GPU allows to run sizeable Deep Learning models without overloading the computational power of the robot.
The D435 camera is connected to this additional GPU to use the 2D/3D images as inputs for deep learning algorithms. The GPU and the D435 camera are powered by the battery of the robot. Currently, the robot provides 795 Wh\footnote{\url{http://doc.aldebaran.com/2-5/family/pepper_technical/battery_pep.html}}. In order to prevent a severe loss in autonomy and respect the Electrical power Constraints, the power consumption of the selected GPU is limited to a maximum consumption of 10\% (79.5 Wh) of the battery of the robot. This constraint narrows the list of candidates to embedded chips that are less power consuming than data center chips and cards\cite{benchmark_gpu}.
The dimension is an important factor for the integration. Indeed to respect the Design Constraints, the new device needs to be integrated while minimizing the impact on the appearance of the robot. In Pepper 1.8, a position behind the tablet in the torso (Fig.\ref{pepper_cooling}) can be used to integrate a small device with a maximum dimension of 140x140x40mm. The Jetson TX2 offers a good trade-off between performance and dimensions for the architecture of the prototype. Moreover, Nvidia provides useful libraries for Deep Learning such as CUDA\footnote{\url{https://developer.nvidia.com/cuda-zone}}, TensorRT\footnote{\url{https://developer.nvidia.com/tensorrt}}, cuDNN\footnote{\url{https://developer.nvidia.com/cudnn}}, etc.
The interface between the Jetson and the robot is done by the Connect Tech Elroy motherboard\footnote{\url{http://connecttech.com/product/asg002-elroy-carrier-for-nvidia-jetson-tx2-tx1}}. This card is designed for the Jetson devices (TX1 and TX2/TX2i) and measures 87x50mm.
The final module (Fig.\ref{fig:module_jetson}) to integrate is composed of:
\begin{itemize}
    \item A power board
    \item An additional SSD memory for storing dataset for Deep Learning algorithms
    \item A carrier board
    \item A Jetson TX2
\end{itemize}{}

%\begin{figure}[htbp]
%\centering
% \includegraphics[scale=0.4]{figs/encastrement.png}
%\caption{Area available behind the tablet of Pepper (unit in mm)}
%label{fig:encastrement_jetson}
%end{figure}

\subsection{Experiments}

\subsubsection{Depth Camera}

\begin{figure}[htbp]
\centering
\subfloat[\label{cam_tripod}]{
 \includegraphics[scale=0.4]{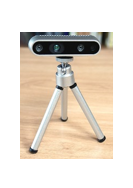}}
\subfloat[\label{cam_pos_a}]{
 \includegraphics[scale=0.4]{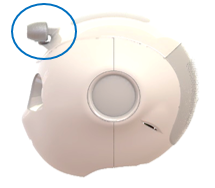}}
 \subfloat[\label{cam_pos_b}]{
 \includegraphics[scale=0.4]{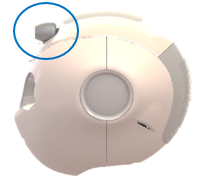}}
\caption{Intel D435 in different configurations: (a) Intel D435 camera on a tripod, (b) Intel D435 camera fixed on the forehead and spaced from the shell, (c) Intel D435 camera fixed on the forehead and built in the shell}
\label{fig:camera_postion_temp}
\end{figure}

The camera is attached to the forehead of Pepper. The main issue is to select the best type of fixation for the camera to avoid overheating. These temperature measures are acquired through thermocouple sensors installed on the dissipator of the camera, in a stabilised environment (25$^{\circ}$C +/- 1$^{\circ}$C), and in 3 different positions (Fig.\ref{fig:camera_postion_temp}).

\begin{figure}[htbp]
\centering
 \includegraphics[scale=0.40]{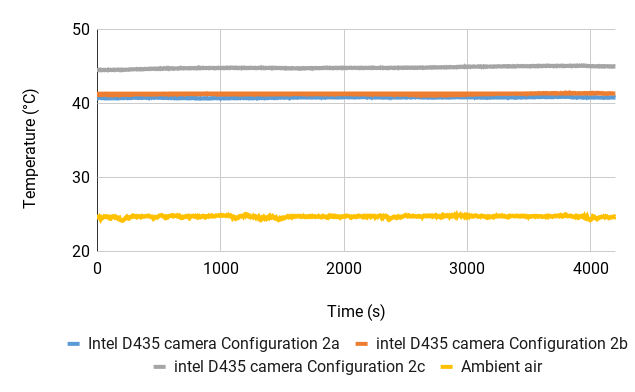}
\caption{ Ambient temperature and heat sink temperature of the camera for three different positions: Intel D435 camera on a tripod (Configuration 2a), Intel D435 camera fixed on the forehead and spaced from the shell (Configuration 2b) and Intel D435 camera fixed on the forehead and built in the shell (Configuration 2c).}
\label{fig:graph_cam_temp}
\end{figure}

To measure the maximum temperature of the camera, the 3D and 2D images are streamed using the official library provided by the constructor\footnote{\url{https://github.com/IntelRealSense/librealsense}} during an hour to increase the temperature of the camera. In order to take into account the heat of the top camera of Pepper, the measurements in the configuration \ref{cam_pos_a} and \ref{cam_pos_b} are done with the robot switched on. The heat sink of the camera is passive, meaning that the dissipation method is based on air convection. The camera must not be enclosed in order for the passive heat sink to be efficient. The measurements confirm this sensitivity with a difference of 5$^{\circ}$C between the configuration \ref{cam_pos_a} and the configuration \ref{cam_pos_b} (Fig.\ref{fig:graph_cam_temp}).

The configuration \ref{cam_tripod} is considered as the optimal position for the 3D camera. In comparison with the configuration \ref{cam_pos_b}, the temperature measured in the configuration \ref{cam_pos_a} is far closer to the configuration \ref{cam_tripod}. This configuration allows an optimal use of the 3D camera and respects the Design Constraints.

\subsubsection{GPU}

First, the heat produced by the Jetson TX2 is characterized with the current drawn at full capacity. To Realize that characterization, the CPU and GPU are used to the maximum of their capacity during an hour. The CPU is stressed with a stress command\footnote{\url{https://manpages.ubuntu.com/manpages/artful/man1/stress-ng.1.html}} and the GPU is stressed by an overload of matrix scalar multiplication with CUDA.

Three different levels of temperature are measured:
\begin{itemize}
    \item Ambient air Temperature: temperature of the ambient air.
    \item Jetson Outer Temperature: temperature on the heat sink of the Jetson.
    \item Jetson Inner Temperature: temperature of the GPU of the Jetson estimated by the software of the Jetson.
\end{itemize}{}

\begin{figure}[htbp]
\centering
\subfloat[\label{graph_tx2_no_fan_temp}]{
 \includegraphics[scale=0.35]{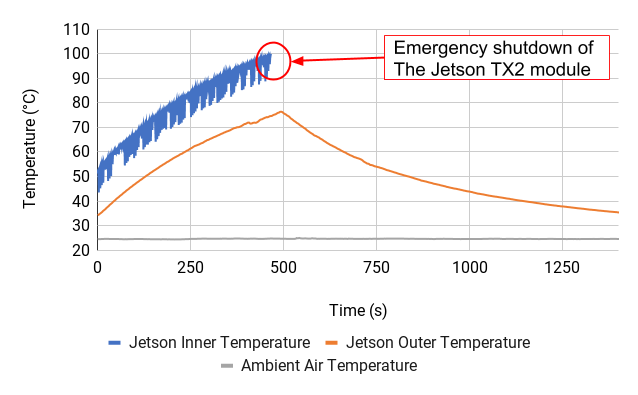}}
 \newline
\subfloat[\label{graph_tx2_fan_temp}]{
 \includegraphics[scale=0.35]{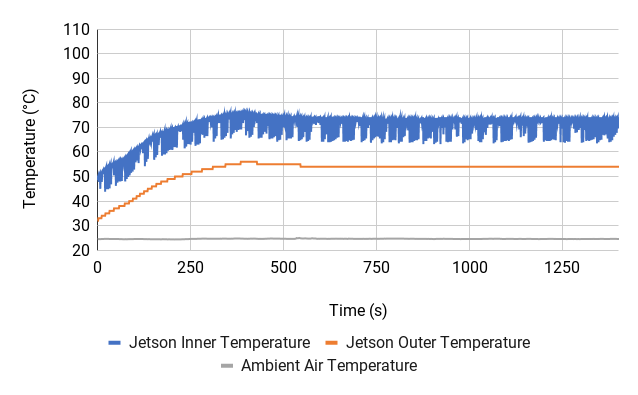}}
\caption{Temperature measurement (Inner/Outer temperature of the Jetson and ambient air temperature) during the stress of the Jetson CPU/GPU with: (a) passive heat sink, (b) active heat sink.}
\label{fig:graph_tx2_temperature_stress}
\end{figure}

The temperatures are captured in two different setups: using a passive dissipator and an active dissipator (fan) or a passive dissipator only to cool the Jetson. With a passive dissipator only, the Jetson Inner Temperature rises to 100$^{\circ}$C in only height minutes (Fig.\ref{graph_tx2_no_fan_temp}). The Jetson safety is engaged and the system shuts down.

%\begin{figure}[htbp]
%\centering
% \includegraphics[scale=0.40]{figs/graph_tx2_no_fan_temp.png}
%\caption{Temperature measurement during the stress of the Jeston %CPU/GPU with passive heat sink}
%\label{fig:graph_tx2_no_fan_temp}
%\end{figure}

As aforementioned, the inner estimated temperature is computed and not measured\footnote{\url{http://developer.nvidia.com/embedded/dlc/jetson-tx2-thermal-design-guide}}. The consequence of this estimation is seen in the Fig.\ref{graph_tx2_no_fan_temp}. These outliers are a byproduct of the GPU stressing script and correspond to brief inactivity periods during the matrix scalar multiplication loop. Indeed, the matrix scalar multiplication script is a loop that overloads the GPU with matrix computation tasks. When the scalar multiplication is over, the program sends another matrix computation to the GPU. The graph exposes the maximum temperature threshold allowed by the processor (100$^{\circ}$C).
The second setup provides a better cooling with the active heat sink: the temperature rises to 80$^{\circ}$C (Fig.\ref{graph_tx2_fan_temp}). Besides, the same gap between the inner estimated and outer temperatures are measured. This gap can be seen in the Fig.\ref{graph_tx2_no_fan_temp} and the Fig.\ref{graph_tx2_fan_temp} (a difference of 20$^{\circ}$C between the outer and inner temperature). In a similar temperature condition, a correlation between the inner and outer temperature can be done in other experiments.

%\begin{figure}[htbp]
%\centering
% \includegraphics[scale=0.4]{figs/graph_tx2_fan_temp.png}
%\caption{Temperature measurement during the stress of the Jeston %CPU/GPU with active heat sink}
%\label{fig:graph_tx2_fan_temp}
%\end{figure}

The active heat sink is able to stabilize the temperature and allows a security margin, preventing the Jetson from reaching the maximum temperature authorized by the system.

\begin{figure}[htbp]
\centering
 \includegraphics[scale=0.4]{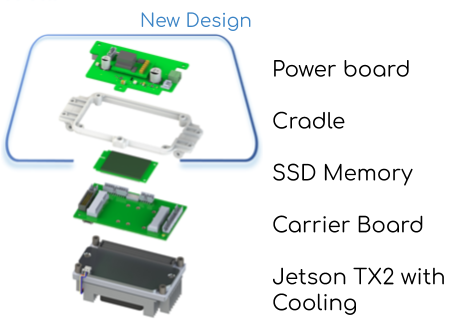}
\caption{Exploded view of the Jetson TX2 module}
\label{fig:module_jetson}
\end{figure}

The current drawn by the complete module (3D Camera and the Jetson Module Fig.\ref{fig:module_jetson}) is also measured while running a stress test. The complete module is powered by a DC current. The estimated maximum power drawn by the complete module does not exceed 30 Wh. The power consumption of the complete module respects the aforementioned constraint of 10\% of the total power provided by the robot (10\% of 795 Wh). Moreover, the Jetson module is composed of a DC/DC converter allowing to adapt the current. The converter has also the capacity to handle twice of the current power to reduce the temperature produced by this component inside the robot.
Finally, a delay block is connected to synchronize the start up of the robot with the one of the complete module. This device avoids safety issues with the robot own battery. Indeed, the battery shuts down dues to a high current draw.
The Fig.\ref{fig:module_jetson} presents an exploded view of the 5 parts that compose the Jetson module with two designed by our team (power board and cradle).

This module is integrated behind the tablet of the robot.

The cooling flow of the Jetson needs to be isolated from the inner part of the pepper to preserve a good thermal exchange.

\begin{figure}[htbp]
\centering
\subfloat[\label{pepper_cooling}]{
 \includegraphics[scale=0.24]{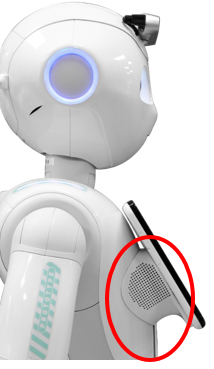}}
\subfloat[\label{cooling_part}]{
 \includegraphics[scale=0.28]{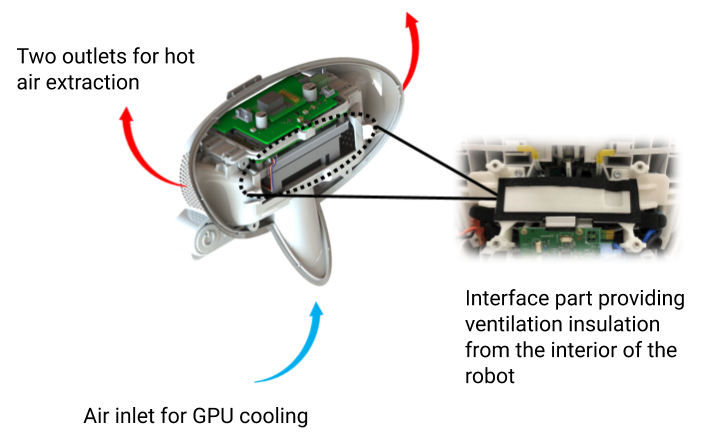}}
\caption{Integration of the module on Pepper: (a) Pepper with the module integrated, (b) Module with the cooling part.}
\label{fig:air_cooling_device}
\end{figure}

The isolation of the cooling flow of the Jetson module from the inside of Pepper is crucial to preserve a good thermal exchange (Fig.\ref{fig:air_cooling_device}).

\begin{figure}[htbp]
\centering
\subfloat[\label{pepper_temp_no_module}]{
 \includegraphics[scale=0.4]{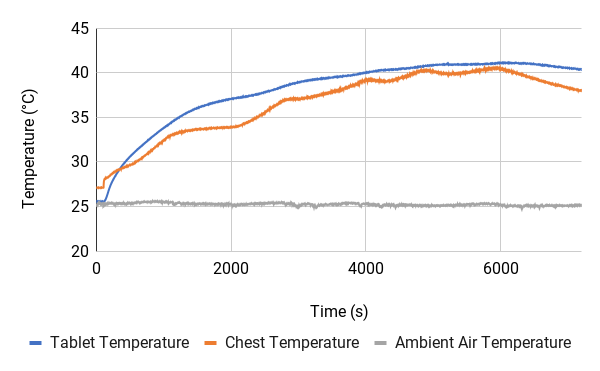}}
\newline
\subfloat[\label{pepper_temp_module}]{
 \includegraphics[scale=0.4]{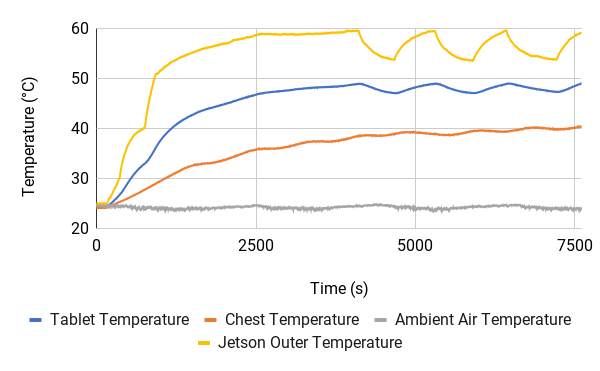}}
\caption{Measure of the temperature of the Tablet, the Chest, the ambient air and the Outer temperature of the Jetson in two different setups: (a) Pepper without the module integrated, (b) Pepper with the module integrated.}
\label{fig:temp_module_jetson_integrated}
\end{figure}

To validate the integration, temperatures are measured in different locations, and in two different setups. First, the temperature of the Pepper 1.8 without the module integrated is measured (Fig.\ref{pepper_temp_no_module}). With the robot turned on, the Tablet Temperature (temperature behind the tablet), the Chest Temperature (Temperature inside the chest of the robot), and Ambient Air Temperature are collected. These values are indicators of a standard temperature in a normal use of the robot. The temperature of the chest and behind the tablet do not exceed 40$^{\circ}$C with an ambient temperature of 25$^{\circ}$C. In the second setup, the new module is integrated and the temperatures measured while stressing the module with the matrix scalar multiplication script. The temperatures described in the first setup (behind the tablet/in the chest/ambient air) are once again acquired. An extra temperature is acquired: the temperature on the Jetson Heat Sink Temperature (Jetson Outer Temperature).

As in the Fig.\ref{pepper_temp_no_module}, the graph of the temperature in the chest (Fig.\ref{pepper_temp_module}) rises to 40$^{\circ}$C. The internal temperature of the robot is not impacted by the integration of the Jetson module. The temperature behind the tablet rises to 46$^{\circ}$C because of the heat produced by the GPU system. As the temperature conditions are similar to the ones in Fig.~\ref{fig:graph_tx2_temperature_stress}, the assumption is to estimate the inner temperature with a difference of 20$^{\circ}$C with the outer temperature. Hence, the internal temperature of the Jetson integrated into Pepper should not exceed 80$^{\circ}$C with an overload of the GPU/CPU. Moreover, the thermal oscillation seen on the graph is due to the activation and deactivation of the active heat sink by the Jetson when the heat of the processor attains 82$^{\circ}$C. The active heat sink is stopped when the temperature drops below 73$^{\circ}$C. Please note that the measures are done in extreme conditions, measures done in nominal conditions would yield lower temperatures.

\subsubsection{Connection between the robot and the devices}

In order to connect the different devices to the robot, an ethernet cable (connection between the head of Pepper and the Jetson) and a USB 3.0 (between the 3D camera and Jetson) are used.

\begin{figure}[htbp]
\centering
 \includegraphics[scale=0.5]{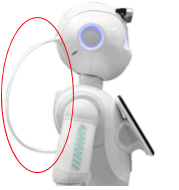}
\caption{Connection between the robot and the devices through the chest with an external white sheath.}
\label{fig:connection_modules}
\end{figure}

Because of the ability of the head of the robot to be easily removable, passing the cable through the neck requires too much time of work. To limit the integration time and respect the integration constraints the choice was made to pass these cables outside of the robot and through the chest instead of the neck. The different modules are connected with an external white sheath and long enough not to disturb the movement of the head of the robot (Fig.\ref{fig:connection_modules}).

To minimize the impact on the design of the robot, the cables pass through the head and the torso instead of a direct connection to the Jeston module.

\section{Software}

Firstly, this section introduces the software architecture of the Adapted Pepper. Secondly, the whole prototype is validated by running Deep Learning models.

\subsection{Architecture}

\begin{figure}[htbp]
\centerline{\includegraphics[scale=0.25]{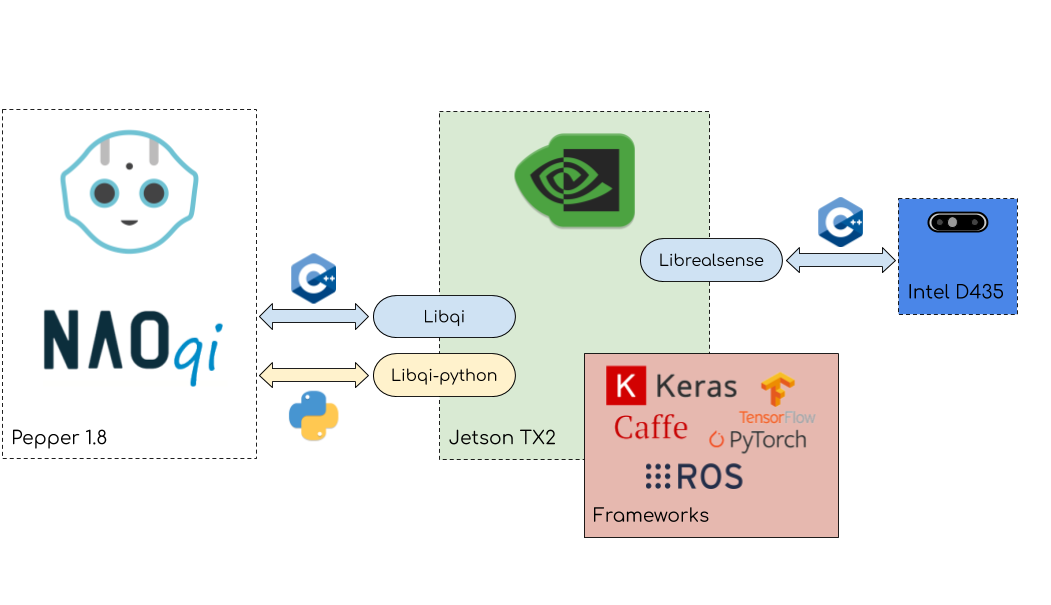}}
\caption{Architecture of the three different elements communicating with each other: Pepper, the Jetson TX2 and the Intel D435 Camera}
\label{fig:jetson_connexion_schema}
\end{figure}

The architecture (Fig.\ref{fig:jetson_connexion_schema}) is composed of three different elements communicating with each other: the robot, the Jetson TX2 (with Jetpack version 3.3 installed\footnote{\url{https://developer.nvidia.com/embedded/jetpack-3_3}}) and the Intel D435 camera. The architecture is based on ROS~\cite{ROS} to interface the different libraries of the three blocks. The Jetson communicates with the camera thanks to the librealsense library and the ROS wrapper, realsense-ros\footnote{\url{https://github.com/IntelRealSense}}.
The communication between the head of the robot and the Jetson is based on a ROS wrapper, naoqi\_driver\footnote{\url{https://github.com/ros-naoqi/naoqi_driver}}, and more precisely on the libqi library\footnote{\url{https://github.com/aldebaran/libqi}}. To use the libqi or libqi-python\footnote{\url{https://github.com/aldebaran/libqi-python}} libraries, a modification of the libraries is needed to enable their compilation for the ARM architecture of the Jetson. Theses libraries allow the connection with NAOqi~\cite{naoqi}, the framework running on the robot.

\subsection{Experiments}

Two different models are tested: OpenPose and YOLO. For validating the architecture and the prototype, the computation of both models needs to be embedded not to overload the network bandwidth and remove network dependency. Moreover, the prototype needs to be faster than a "standard" Pepper using external computation.

\subsubsection{OpenPose}

OpenPose provides different models: MPII, COCO, BODY\_25. Only the COCO and BODY\_25 models are compared in this section. To optimize the OpenPose inference, the Caffe model file of a trained model is converted into a GIE (Nvidia GPU Inference Engine) object more adapted to the GPU of the Jetson, by using TensorRT (TRT). This operation is possible for the COCO model but some issues are encountered with the BODY\_25 model. Indeed, the TRT library (version 4.0) does not support PReLU layers required to run this OpenPose model. A solution to support it would be to create a TRT plugin. Experiments show that this method breaks the CBR (combination of Convolution, Bias and activation Relu) process in TRT and is very slow. Another workaround is to replace each PReLu operation by a combination of Relu layer, Scale layer and ElementWise addition Layer. However, the caffe model needs to be trained with this modified version of BODY\_25 model. The training can be burdensome and there is no confidence about the final accuracy or speed of this new model. In a future version, an upgrade of Jetpack to Jetpack 4.3 that support PReLU, thanks to TRT 6.0, is considered.
The accuracy and the framerate obtained by the three different models are compared: COCO (caffe and TRT model) and BODY\_25 (caffe model). Because of the computational power limitation of the device, it was not possible to run OpenPose with a high input resolution. Indeed, the input net resolution must be reduced below 256x256 to get a framerate above 5 FPS (Tab.\ref{tab:openpose_model_fps}). The input net resolution needs to be reduced even more for the COCO model. The inference speed of the COCO model has been improved a lot by TRT (1.8 times faster) and is comparable to the inference speed of the BODY\_25 model. 

\begin{center}
\begin{table}[htbp]
    \centering
    \caption{Number of frames per seconds of three OpenPose models running on the Jetson TX2}
    \label{tab:openpose_model_fps}
    \begin{tabular}{|c|c|c|c|c|}
        \hline
        input & BODY\_25 & COCO & COCO \\
        net resolution & (Caffe) & (Caffe) & (TRT)\\
        \hline
        368x368 & 2.4 & 0.9 & 1.7\\
        \hline
        256x256 & 5.1 & 1.8 & 5.7\\
        \hline
        256x128 & 8.9 & 6.6 & 8.6\\
        \hline
        128x128 & 15.7 & 8.5 & 13.6\\
        \hline
    \end{tabular}
    
\end{table}
\end{center}

\begin{figure}[htbp]
\centering
\subfloat[\label{average_precision}]{
 \includegraphics[scale=0.4]{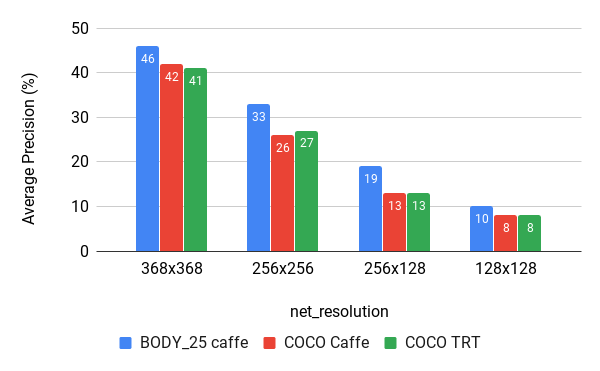}}
 \newline
\subfloat[\label{average_recall}]{
 \includegraphics[scale=0.4]{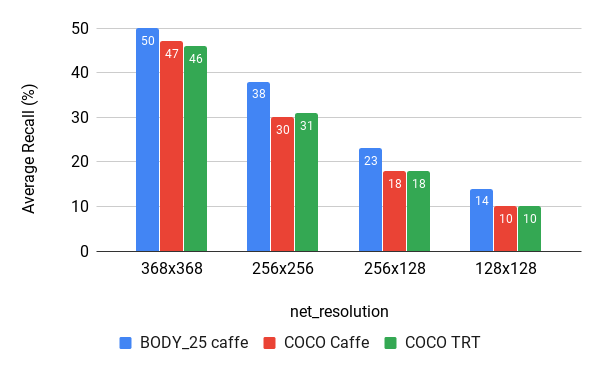}}
\caption{Metrics on the COCO validation set (2017): (a) Average Precision (at human keypoints similarity=.50:.05:.95) of the different models, (b) Average Recall (at human keypoints similarity=.50:.05:.95) on human keypoints of the different models}
\label{fig:ap_ar_openpose}
\end{figure}

For each model, the Average Precision (AP) and Average Recall (AR) on the COCO~\cite{COCO_dataset} validation set (2017) are tested. Whereas both COCO model obtain as expected a similar score, the accuracy of BODY\_25 is better for each resolution (Fig.\ref{fig:ap_ar_openpose}). However, in order to obtain the maximum AP as in \cite{openpose}(61.8\%), the net resolution needs to be above 368x368.

In order to preserve an optimal accuracy and framerate with the use of another neural network, the net resolution of 256x256 with the BODY\_25 model can be an appropriate compromise.

\subsubsection{OpenPose and YOLO}

The usage of YOLO and OpenPose in parallel is tested to simulate a use case of the aforementioned project MuMMER, head pose estimation from OpenPose and object detection. The inference speed of tiny YOLO v3 is already fast enough on the Jetson (25.0 FPS with a 2D image resolution of 640x480 pixels and an input net resolution of 416x416). The YOLO model is adapted to interface the model with the input of the Intel D435 camera\footnote{\url{https://github.com/softbankrobotics-research/darknet_ros}}. Three ROS nodes are launched in parallel:
\begin{itemize}
    \item realsense (resolution color camera at 648x480 pixels)
    \item OpenPose (input net\_resolution of 256x256 and BODY\_25 model)
    \item tiny YOLO v3 (input net\_resolution of 416x416)
\end{itemize}{}
As a consequence of running two models in parallel, their respective framerate (Tab.\ref{tab:models_fps_test}) is divided by two. The framerate of OpenPose dropped from 5.1 to 2.6 FPS. The bottleneck of the architecture is limited by the embedded computational power and not by the bandwidth of the network interface.

\begin{center}
\begin{table}[htbp]
    \centering
    \caption{Framerate of two models running in parallel on the Jetson TX2}
    \label{tab:models_fps_test}
    \begin{tabular}{|c|c|}
        \hline
        Model & Framerate (Image per seconds) \\
        \hline
        OpenPose & 2.6 \\
        \hline
        Tiny YOLO V3 & 11.6\\
        \hline
    \end{tabular}
    
\end{table}
\end{center}

\subsubsection{Embedded vs non-Embedded}

The performances of the Adapted Pepper (embedded computation) and the performances of a "standard" Pepper 1.8 (non embedded computation) are compared. The images from the top camera of Pepper 1.8 are sent through WiFi to an external computer. The external computer embeds a 1080 Tegra Ti GPU. The images retrieved by the external computer with naoqi driver (resolution of 640x480 pixels) are used as input for the YOLO model (tiny YOLO v3, input net\_resolution of 416x416). This setup reaches a framerate of 10.3 images per second. The embedded computation obtains a much better framerate, 25.0 FPS. The non embedded computation performance is due to the WiFi bandwidth. Indeed, the images are received through the WiFi at a frequency of 11.4 FPS. This bottleneck is not observed with the Jetson TX2 directly connected to the Intel D435 camera. The bottleneck due to the network interface is highlighted with the stereo image of Pepper 1.8. Whereas the framerate obtained for each 3D resolution with the Adapted Pepper and the D435 camera are 30 FPS, the framerate obtained with the stereo camera of Pepper 1.8 collected through WiFi reaches 15 FPS at a requested resolution of 320x180 pixels (Tab.\ref{tab:3d_fps_pepper}).

\begin{center}
\begin{table}[htbp]
    \centering
    \caption{Framerate of the stereo image of Pepper 1.8 through WiFi}
    \label{tab:3d_fps_pepper}
    \begin{tabular}{|c|c|}
        \hline
        camera resolution (pixels) & Framerate (image per seconds)\\
        \hline
        1280x720 & 1.6 \\
        \hline
        640x360 & 4.8\\
        \hline
        320x180 & 14.8\\
        \hline
    \end{tabular}
    
\end{table}
\end{center}

\section{CONCLUSION}

This paper presents a prototype called Adapted Pepper embedding a GPU (Jetson TX2) and an additional camera (intel realsense D435). These hardware modifications increase the computational power and the sensing capabilities of the robot, allowing to run Deep Learning algorithms in an embedded fashion, hence improving the capabilities of Pepper. By removing the network factors, the robot is not network dependent anymore, more secured, and the bottleneck of the solution is the embedded GPU capabilities, which is a controlled factor. However, the total computational power available on the robot is limited compared to graphic cards for computers (like the one used in experiments). Efforts of integration and optimization were needed to optimize the inference speed. To minimize the bandwidth usage, preprocessing the sensors data with embedded computation reduces the amount of data to be sent to an external processing unit. The prototype can be improved by using the latest version of Jetpack offering an update on essential libraries (CUDA, TensorRT, cuDNN, etc.). Replacing the Jetson TX2 with a device with more computational power could also be envisioned to improve the overall performances.

\bibliographystyle{unsrt}
\bibliography{biblio}

\newpage

\appendix

\subsection{Camera Integration}

\begin{figure}[htbp]
\centering
\subfloat[\label{adapted_pepper_inside_head}]{
 \includegraphics[scale=0.27]{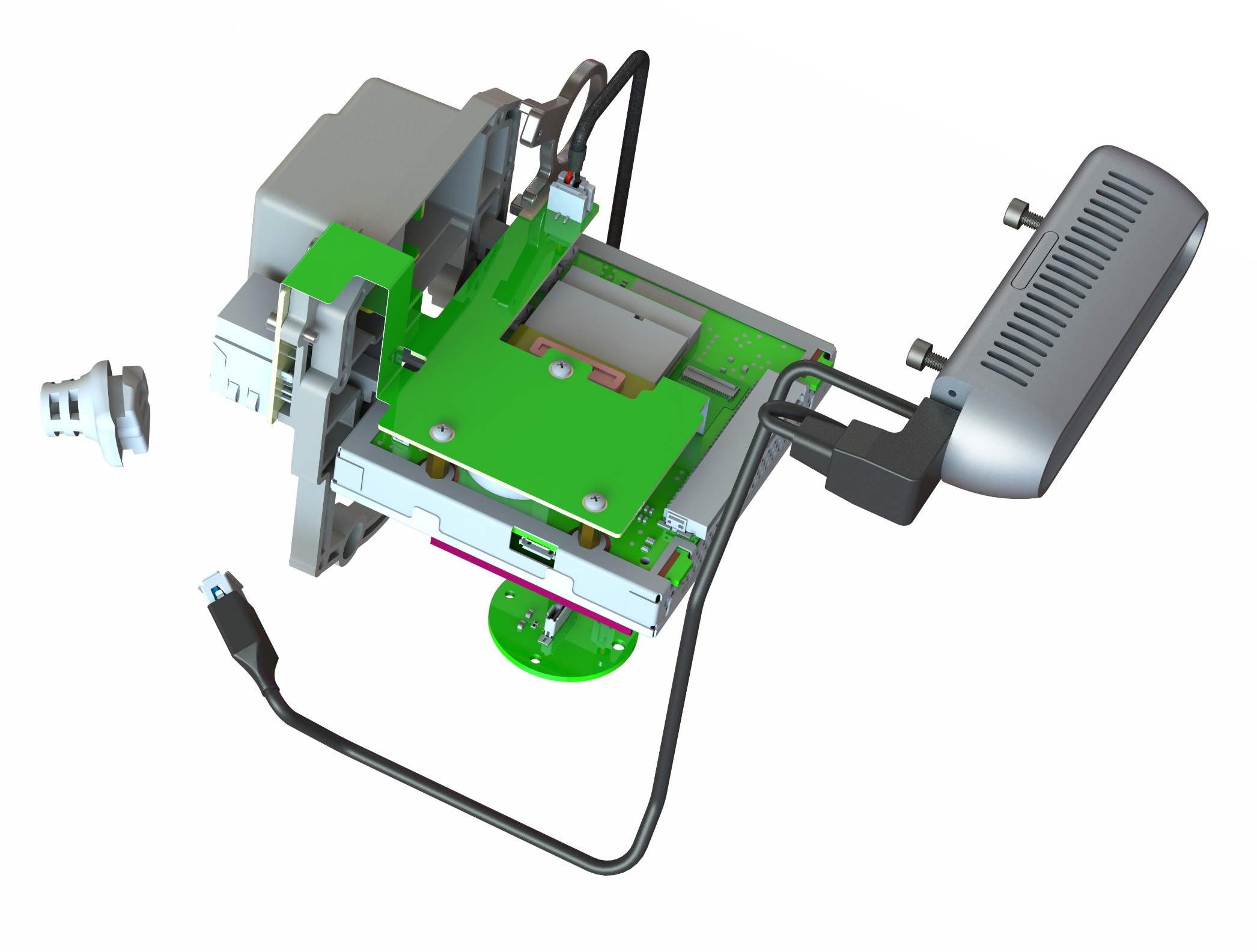}}
\subfloat[\label{Adapted_Pepper_head}]{
 \includegraphics[scale=0.25]{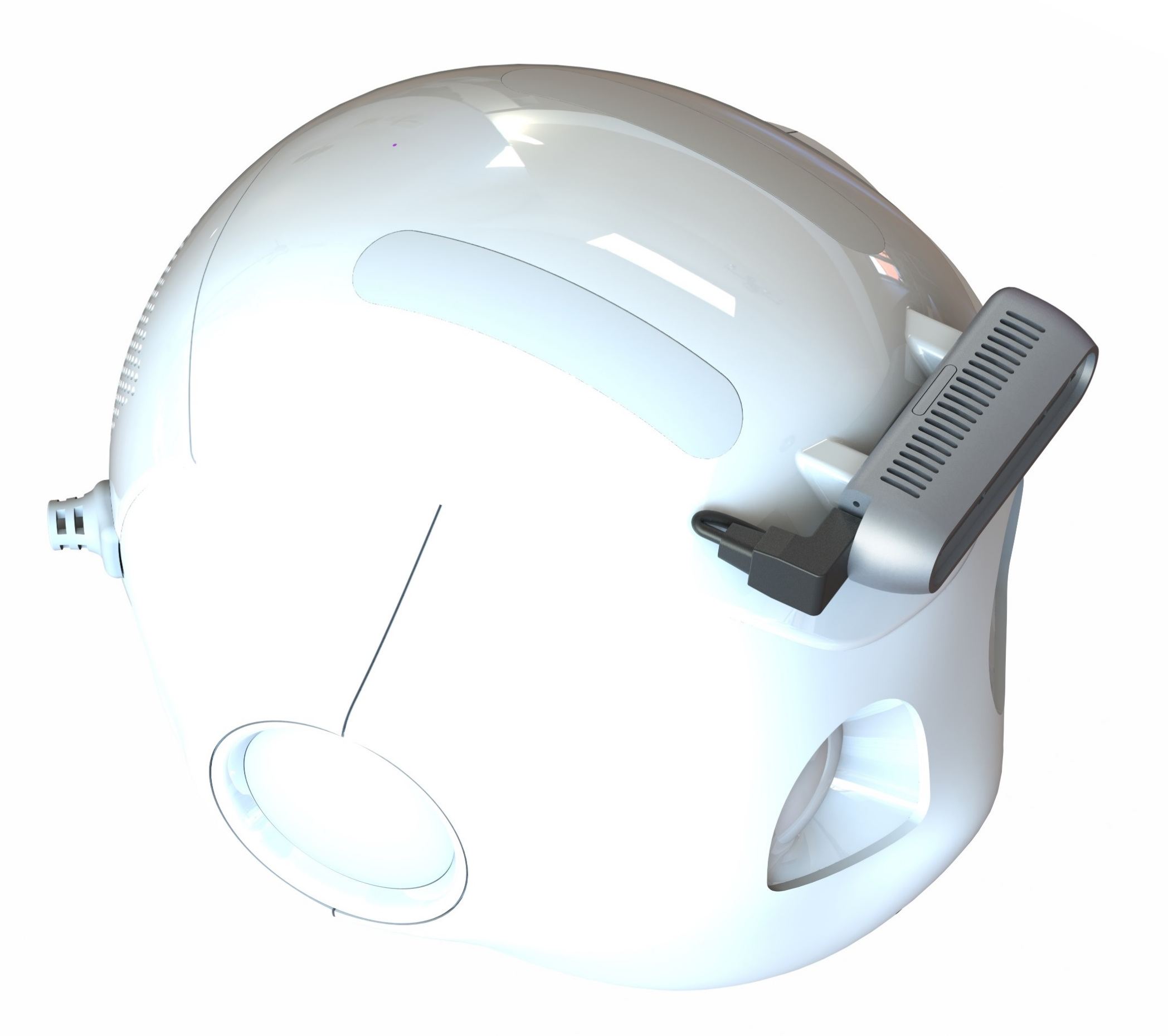}}
\caption{Camera integration: (a) Exploded view of the camera integration, (b) Integrated result of the camera integration}
\label{fig:Camera_integration}
\end{figure}

\subsection{Jetson TX2 module Integration}

\begin{figure}[htbp]
\centering
 \includegraphics[scale=0.5]{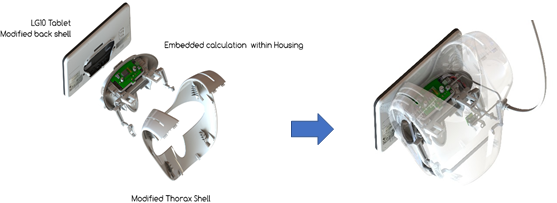}
\caption{Integration of the Jetson TX2 module inside the torso of Pepper}
\label{fig:jetson_integration_torso}
\end{figure}

\end{document}